\documentclass{article} 
\usepackage{iclr2016_conference,times}
\usepackage{hyperref}
\usepackage{url}
\usepackage{graphicx}
\usepackage{xspace}
\usepackage{caption}
\usepackage{amsmath} 
\usepackage{amssymb}

\usepackage{color}

\definecolor{dgreen}{rgb}{0.0,0.4,0.0}
\definecolor{dred}{rgb}{1,0.0,0.0}

\title{Evaluating Prerequisite Qualities for Learning End-to-End Dialog Systems}

\author{Jesse Dodge\thanks{The first three authors contributed equally.},~ Andreea Gane$^{*}$,~ Xiang Zhang$^{*}$, Antoine Bordes, \\
 {\bf Sumit Chopra, ~Alexander H. Miller, ~Arthur Szlam \& Jason Weston }\\
Facebook AI Research\\
770 Broadway\\
New York, USA\\
\texttt{\{jessedodge,agane,xiangz,abordes,spchopra,ahm,aszlam,jase\}@fb.com}
}

\iclrfinalcopy 

\begin{document}

\maketitle

\begin{abstract}
A long-term goal of machine learning is to build intelligent conversational agents.
One recent popular approach is to train end-to-end models
on a large amount of real dialog transcripts between humans \citep{sordoni2015neural,vinyals2015neural,shang2015neural}.
However, this approach leaves many questions unanswered as an understanding of the precise 
successes and shortcomings of each model is hard to assess.
A contrasting recent proposal are the bAbI tasks \citep{weston2015towards}
which are synthetic data that measure the ability of learning
machines at various reasoning tasks over toy language.
Unfortunately, those tests are very small and hence may encourage methods that do not scale.
In this work, we propose a suite of new tasks of a much larger scale that attempt to bridge
the gap between the two regimes. Choosing the domain of movies, we provide tasks that test the ability of models to answer factual questions (utilizing OMDB), provide personalization (utilizing MovieLens),
carry short conversations about the two, and finally to perform on natural dialogs from Reddit.
We provide a dataset covering  $\sim$75k movie entities and with $\sim$3.5M training examples.
We present results of various models on these tasks, and evaluate their performance.
\end{abstract}

\section{Introduction}

With the recent employment of Recurrent Neural Networks (RNNs) and the large quantities of conversational data available on websites like Twitter or Reddit, a new type of dialog system is emerging. Such \emph{end-to-end dialog systems} \citep{ritter2011data,shang2015neural,vinyals2015neural,sordoni2015neural} directly generate a response given the last user utterance and (potentially) the context from previous dialog turns without relying on the intermediate use of a dialog state tracking component like in traditional dialog systems (e.g. in \citet{henderson2015machine}). 
These methods are trained to imitate user-user conversations and do not need any hand-coding of attributes and labels for dialog states and goals like state tracking methods do. Being trained on large corpora, they are robust to many language variations and seem to mimic human conversations to some extent.

In spite of their flexibility and representational power, these neural network based methods lack pertinent goal-oriented frameworks to validate their performance.
Indeed, traditional systems have a wide range of well defined evaluation paradigms and benchmarks
that measure
  their ability to track user states and/or to reach user-defined goals \citep{walker1997paradise,paek2001empirical,griol2008statistical,williams2013dialog}. Recent end-to-end models, on the other hand, rely either on very few human scores \citep{vinyals2015neural}, crowdsourcing \citep{ritter2011data,shang2015neural} or machine translation metrics like BLEU \citep{sordoni2015neural} to judge the quality of the generated language only.
This is problematic because these evaluations do not assess if end-to-end systems can conduct dialog to achieve pre-defined objectives, but simply whether they can generate correct language that could fit in the context of the dialog; in other words, they quantify their \emph{chit-chatting} abilities.

To fill in this gap, this paper proposes a collection of four tasks designed to evaluate different pre-requisite qualities of end-to-end dialog systems.
Focusing on the movie domain, we propose to test if systems are able to jointly perform: (1)  question-answering (QA), (2) recommendation, (3) a mix of recommendation and QA and (4) general dialog about the topic, which we call {\em chit-chat}.
All four tasks have been chosen because they test basic capabilities we expect a dialog system performing insightful movie recommendation should have while evaluation
on each of them can be well defined without the need of human-in-the-loop (e.g. via Wizard-of-Oz strategies \citep{whittaker2002fish}).
Our ultimate goal is to validate if a single model can solve the four tasks at once, which we 
assert 
is a pre-requisite for an end-to-end dialog system supposed to act as a movie recommendation assistant,
and by extension a general dialog agent as well.
At the same time we advocate developing methods that make no special engineering for this domain,
and hence should generalize to learning on tasks and data from other domains easily.

In contrast to the bAbI tasks which test basic capabilities of story understanding systems \citep{weston2015towards}, the tasks have been created using large-scale real-world sources (OMDb\footnote{{\tt http://en.omdb.org}}, MovieLens\footnote{{\tt http://movielens.org}} and Reddit\footnote{{\tt http://reddit.com/r/movie}}). 
%
Overall, the dataset covers  $\sim$75k movie entities (movie, actor, director, genre, etc.) with $\sim$3.5M training examples: even if the dataset is restricted to a single domain, it is large and allows a great variety of discussions, language and user goals.
We evaluate on these tasks the performance of various neural network models that can potentially create end-to-end dialogs, ranging from simple supervised embedding models \citep{bai2009supervised},
RNNs with Long Short-Term Memory (LSTMs) \citep{hochreiter1997long}, and attention-based models, in 
particular Memory Networks \citep{sukhbaatar2015end}.
To validate the quality of our results, we also apply our best
performing model, Memory Networks, in other conditions by comparing it on the Ubuntu Dialog Corpus
\citep{lowe2015ubuntu} against baselines trained by the authors of the corpus. We show that they outperform all baselines by a wide margin.

\section{The Movie Dialog Dataset}

We introduce a set of four tasks to test the ability of end-to-end dialog systems,
focusing on the domain of movies and movie related entities.
They aim to test five abilities which we postulate as being key towards a
fully functional general dialog system (i.e., not specific to movies per se):
\begin{itemize}
\item {\bf QA Dataset:} Tests the ability to answer factoid questions that can be answered without relation to previous dialog. The context consists of the question only. 
\item {\bf Recommendation Dataset:} Tests the ability to provide personalized responses to the user via recommendations (in this case, of movies) rather than universal facts as above. 
\item {\bf QA+Recommendation Dataset:} Tests the ability of maintaining short dialogs involving both factoid and personalized content where conversational state has to be maintained.
\item {\bf Reddit Dataset:} Tests the ability to identify most likely replies in discussions on Reddit.
\item {\bf Joint Dataset:} All our tasks are dialogs. They can be combined into a single dataset,
testing the ability of an end-to-end model to perform well at all skills at once.
\end{itemize}

Sample input contexts and target replies from the tasks are given in Tables
\ref{taskset1}-\ref{taskset4}.
The datasets are available at: \url{http://fb.ai/babi}.

\subsection{Question Answering (QA)} \label{sec-qa}

The first task we build is to test whether a dialog agent is capable of answering simple
factual questions. 
The dataset was built from the Open Movie Database (OMDb)\footnote{Downloaded from \url{http://beforethecode.com/projects/omdb/download.aspx}.} 
which contains  metadata about movies.
The subset we consider contains $\sim$15k movies, $\sim$10k actors and $\sim$6k directors.
We also matched these movies to the MovieLens dataset\footnote{\url{http://grouplens.org/datasets/movielens/}} to attribute tags to each movie.
We  build a knowledge base (KB) directly from the combined data, stored as triples such
as {\sc (The Dark Horse, starred\_actor, Bette Davis)} and
{\sc (Moonraker, has\_tag, james bond)}, with 8 different relation types
involving director, writer, actor, release date, genre, tags, rating and imdb votes.


We distinguish 11 classes of question, 
corresponding to different kinds of edges in our KB:
{\em actor to movie} (``What movies did Michael J Fox star in?''),
{\em movie to actors} (``Who starred in Back to The Future?''),
{\em movie to director}, {\em director to movie},
{\em movie to writer}, {\em writer to movie},
{\em movie to tags}, {\em tag to movie},
{\em movie to year}, {\em movie to genre} and {\em movie to language}.
For each question type there is a set of possible answers.
Using SimpleQuestions, an existing open-domain question answering dataset based on Freebase 
\citep{bordes2015large} we identified
 the subset of questions posed by those human annotators that covered
our question types. We expanded this set to cover all of our KB by 
substituting the actual entities in those questions to also
apply them to other questions, e.g. if the original question written by an annotator was
``What movies did Michael J Fox star in?'', we created a pattern
``What movies did [@actor] star in?'' which we substitute for any
other actors in our set, and repeat this for all annotations.
We split the questions into training, development and test sets 
with $\sim$96k, 10k and 10k examples, respectively.

\begin{table*}[h!]
\begin{small}
\begin{center}
\begin{tabular}{|l|}
\hline
{\bf Task 1: Factoid Question Answering (QA)}    \\
 \\[-1ex]
~~What movies are about open source?   \textcolor{dred}{Revolution OS}\\
~~Ruggero Raimondi appears in which movies?     \textcolor{dred}{Carmen}\\
~~What movies did Darren McGavin star in?       \textcolor{dred}{Billy Madison, The Night Stalker, Mrs. Pollifax-Spy}\\
~~Can you name a film directed by Stuart Ortiz? \textcolor{dred}{Grave Encounters}\\
~~Who directed the film White Elephant?  \textcolor{dred}{Pablo Trapero}\\
~~What is the genre of the film Dial M for Murder?   \textcolor{dred}{Thriller, Crime}\\
~~What language is Whity in?     \textcolor{dred}{German}\\
\hline
\end{tabular}
\caption{Sample input contexts and target replies (in red) from Task 1.  \label{taskset1}}\vspace{2mm}
\end{center}
\end{small}
\vspace*{-3ex}
\end{table*}

To simplify evaluation rather than requiring the generation of sentences containing the answers,
we simply ask a model to output a list, which is ranked as the
possible set of answers. We then use standard ranking metrics to
evaluate the list, making the results easy to interpret. Our main results report  the 
hits@1 metric (i.e. is the top answer correct); other metrics are given in the appendix.

\subsection{Recommendation Dataset}

Not all questions about movies in dialogs have an objective answer, independent of the
person asking; indeed much of human dialog is based on opinons and personalized responses.
One of the simplest dialogs of this type to evaluate is that of recommendation, 
where we can utilize existing data resources.
We again employ the MovieLens dataset
which features a user $\times$ item matrix
of movie ratings, rated from 1 to 5. We filtered the set of movies to be the same set as in
the QA task and additionally only kept movies that had at least 2 ratings, giving 
around $\sim$ 11k movies.

To use this data for evaluating dialog, 
we then use it to generate dialog exchanges.
We first select a user at random; this will be the user who is participating in the dialog,
and then sample 1-8 movies that the user has rated 5. We then form a statement intended
to express the user's feelings about these movies,
 according to a fixed set of natural language templates, one of which is selected randomly.
See Table \ref{taskset2} for some examples.
From the remaining set of movies the same user gave a rating of 5, we select one to be the answer.


\begin{table*}[h]
\begin{center}
\begin{small}
\begin{tabular}{|l|}
\hline
{\bf Task 2: Recommendation}    \\
 \\[-1ex]
Schindler's List, The Fugitive, Apocalypse Now, Pulp Fiction, and The Godfather are films I really liked. \\Can you suggest a film?         \textcolor{dred}{The Hunt for Red October}\\
 \\[-1ex]
Some movies I like are Heat, Kids, Fight Club, Shaun of the Dead, The Avengers, Skyfall, and Jurassic Park. \\Can you suggest something else I might like?    \textcolor{dred}{Ocean's Eleven}\\
\hline
\end{tabular}
\caption{Sample input contexts and target replies (in red) from Task 2.  \label{taskset2}}\vspace{2mm}
\end{small}
\end{center}
\vspace*{-3ex}
\end{table*}

There are $\sim$110k users in the training, $\sim$1k users in the development set
and  $\sim$1k  for test.
We follow the procedure above sampling users with replacement 
and generate 1M training examples and 10k development
 and test set examples, respectively.
To evaluate the performance of a model, 
just as in the first task, we evaluate a ranked list of answers.
In our main results we measure hits@100, i.e.  1 if the provided
answer is in the top 100, and 0 otherwise, rather than hits@1 as this task is harder than the last.

Note that we expect absolute hits@k numbers to be lower for this task
than for QA due to incomplete labeling (``missing ratings''):
 in recommendation there is no exact right answer,
and it is not surprising the actual single true label is not always at the top position, i.e. the top predictions of
the model may be good as well, but we do not have their labels.
One can thus view the ranking metric as a kind of lower bound on performance of actually labeling all the predictions using human annotations, which would be time consuming and no longer automatic, and hence undesirable for algorithm development.
This is standard in recommendation, see e.g. \cite{cremonesi2010performance}.

\subsection{QA+Recommendation Dialog}

The tasks presented so far only involve questions followed by responses,
with no context from previous dialog. 
This task aims at evaluating responses in the context of multiple previous exchanges,
while remaining straightforward enough  that evaluation and analysis are still tractable.
We hence combine the question answering and recommendation tasks from before in 
a multi-response dialog, where dialogs consist of 3 exchanges (3 turns from each participant).

The first exchange requires a {\em recommendation} similar to Task 1
 except that they also specify what genre or topic they are interested in,
e.g. ``I'm looking for a Music movie'', where the answer might be
``School of Rock'', as in the example of Table \ref{taskset3}.

In the second exchange, given the model's response (movie suggestion), the user asks a factoid question 
about that suggestion, e.g. ``What else is that about?'', ``Who stars in that?'' and so on.
This question refer back to the previous dialog, making context important.

In the third exchange, the user asks for a alternative
recommendation, and provides  extra information about their tastes, e.g.
 ``I like Tim Burton movies more''. 
Again, context of the last two exchanges should help for best performance.

\begin{table*}[h!]
\begin{center}
\begin{small}
\begin{tabular}{|l|}
\hline
{\bf Task 3: QA + Recommendation Dialog}    \\
 \\[-1ex]
I loved Billy Madison, My Neighbor Totoro, Blades of Glory, Bio-Dome, Clue, and Happy Gilmore.\\
I'm looking for a Music movie.      \textcolor{dred}{School of Rock}\\
What else is that about?       \textcolor{dred}{Music, Musical, Jack Black, school, teacher, Richard Linklater, rock, guitar}\\
I like rock and roll movies more. Do you know anything else?   \textcolor{dred}{Little Richard}\\
\\[-1ex]
Tombstone, Legends of the Fall, Braveheart, The Net, Outbreak, and French Kiss are films I really liked. \\
I'm looking for a Fantasy movie.    \textcolor{dred}{Jumanji}  \\
Who directed that?      \textcolor{dred}{Joe Johnston} \\
I like Tim Burton movies more. Do you know anything else?      \textcolor{dred}{Big Fish}\\
\hline
\end{tabular}
\caption{Sample input contexts and target replies (in red) from Task 3. \label{taskset3}}\vspace{2mm}
\end{small}
\end{center}
\vspace*{-3ex}
\end{table*}

We thus generate 1M examples of such 6 line dialogs (3 turns from each participant) for training,
and $\sim$10k for development and testing respectively.
We can evaluate the performance of models across all the lines of dialog
(e.g., all $\sim$30k responses from the test set), but also only on the 1st (Recommendation), 2nd (QA) 
or 3rd exchange (Similarity) for a more fine-grained analysis.
We again use a ranking metric (here, hits@10), just as in our previous tasks.

\subsection{Reddit Discussion}


Our fourth task is to predict responses in movie discussions using
real conversation data taken directly from Reddit,
 a website where registered community members can submit content in various areas of interest,
 called ``subreddits''.
We selected the movie subreddit\footnote{\url{https://www.reddit.com/r/movies}, selecting from the dataset available at \url{https://www.reddit.com/r/datasets/comments/3bxlg7.}} to match our other tasks.

The original discussion data is potentially between multiple participants.
To simplify the setup, we flatten this to appear as two participants (parent and comment),
just as in our other
tasks. In this way we collected $\sim$1M dialogs, of which 10k are reserved for a development set,
and another 10k for the test set.
Of the dialogs, $\sim$76\% involve a single exchange,
 $\sim$17\% have at least two exchanges, and 7\% have at least three exchanges (the longest exchange is length 50).

\begin{table*}[h!]
\begin{small}
\begin{center}
\begin{tabular}{|l|}
\hline
{\bf Task 4: Reddit Discussion}    \\
 \\[-1ex]
I think the Terminator movies really suck, I mean the first one was kinda ok, but after that they got really \\ cheesy. Even the second one which people somehow think is great. And after that... forgeddabotit.\\
\textcolor{dred}{C'mon the second one was still pretty cool.. Arny was still so badass, as was Sararah Connor's character..}\\
\textcolor{dred}{and the way they blended real action and effects was perhaps the last of its kind...} \\
\hline
\end{tabular}
\caption{Sample input contexts and target replies (in red) from Task 4.  \label{taskset4}}\vspace{2mm}
\end{center}
\end{small}
\vspace*{-3ex}
\end{table*}

To evaluate the performance of models, 
we again separate the problem of evaluating the quality of a response
from that of language generation by considering a ranking setup, in line 
with other recent works \citep{sordoni2015neural}.
We proceed as follows: we select a further 10k 
comments for the development set and another 10k for the test set which have not appeared elsewhere in
 our dataset,  
and use these as potential candidates for ranking during evaluation.
For each exchange, given the input context, we rank 10001 possible candidates:
the true response given in the dataset, plus the 10k ``negative'' candidates just described.
The model has to rank the true response as high as possible.
Similar to recommendation as described before we do not expect absolute hits@k performance to be as 
high as for QA due to incomplete labeling.
As with Task 3, we can evaluate on all the data, or only on the 1st, 2nd or 3rd exchange, and so on.
We also identified the subset of the test set where there is an entity match with 
at least two entities from Tasks 1-3, where one of the entities appears in the input, and
the other in the response: this subset serves to evaluate the
impact of using a knowledge base for conducting such a dialog.

\subsection{Joint Task}

Finally, we consider a task made of the combination of all four of
the previous ones. 
At both training and test time examples consist of exchanges from any of the datasets,
sampled at random, whereby the conversation is
`reset' at each sample, so that the context history only
ever includes exchanges from the current conversation.

We consider this to be the most important task, as it tests whether a model can not only produce {\em chit-chat}
(Task 4) but also can provide meaningful answers during dialog (Tasks 1-3). On the other hand, the point of
delineating the separate tasks is to evaluate exactly which types of dialog a model is succeeding at or not.
That all the datasets are in the same domain is crucial to testing the ability of models at performing well 
on all tasks jointly. If the domains were different, then the vocabularies would be trivially non-overlapping,
allowing to learn effectively separate models inside a single one.

\subsection{Relation to Existing Evaluation Frameworks}

Traditional dialog systems consist of two main modules:
(1) a dialog state tracking component that tracks what has happened in a dialog, incorporating into a pre-defined explicit state structure system outputs, user utterances, context from previous turns, and other external information, and (2) a response generator.
Evaluation of the dialog state tracking stage is well defined since the PARADISE framework \citep{walker1997paradise} and subsequent initiatives \citep{paek2001empirical,griol2008statistical}, including recent competitons \citep{williams2013dialog,henderson2014second} as well as situated variants \citep{rojas2012end}. However, they require fine grained data annotations in terms of labeling internal dialog state and precisely defined user intent (goals). As a result, they do not really scale to large domains and dialogs with high variability in terms of language.
Because of language ambiguity and variation, evaluation of the response generation step is complicated and usually relies on human judgement \citep{walker2003trainable}.

End-to-end dialog systems do not rely on explicit internal state and hence do not have state tracking modules, they directly generate responses given user utterances and dialog context and hence can not be evaluated using state tracking test-beds.
Unfortunately, as for response generator modules, their evaluation is ill-defined as it is difficult to objectively rate at scale the fit of returned responses. Most existing work \citep{ritter2011data,shang2015neural,vinyals2015neural,sordoni2015neural} chose to use human ratings,
which does not easily scale.
\cite{sordoni2015neural} also use the BLEU score to compare to actual user utterances but
this is not a completely satisfying measure of success,
especially when used in a {\em chit-chat} setting where there are no clear goals and hence
measures of success.
%
\cite{lowe2015ubuntu} use a similar ranking evaluation to ours, but only in a {\em chit-chat} setting.

Our approach of providing a collection of tasks to be jointly solved is related to the evaluation framework of the bAbI tasks \citep{Weston15} and of the collection of sequence prediction tasks of \citet{Joulin15}. However, unlike them, our Tasks 1-3
 are much closer to real dialog, 
being built from human-written text,
and with Task 4 actually involving real dialog from Reddit.
%
The design of our tasks is such  that all test one or more key characteristics a dialog system should have but also that an unambiguous answer is expected after each dialog act. In that sense, it follows the the notion of dialog evaluation by a reference answer introduced in \citep{hirschman1990beyond}.
The application of movie recommender systems is connected to that of TV program suggestion proposed by \cite{ramachandran2014end}, except that we frame it so that we can generate systematic evaluation from it, where they only rely on human judgement at small scale.

\section{Models}


\subsection{Memory Networks}

Memory Networks \citep{weston2014memory,sukhbaatar2015end} are a recent class of models that
 perform language understanding by incorporaring a memory component that potentially
includes both long-term memory (e.g., to remember facts about the world) and short-term context
(e.g., the last few turns of dialog). They have only been evaluated in a few setups:
question answering \citep{bordes2015large}, language modeling \citep{sukhbaatar2015end,hill2015goldilocks},
and language understanding on the 
bAbI tasks \citep{Weston15}, but not so far on dialog
tasks such as ours.

We employ the MemN2N architecture of \citet{sukhbaatar2015end} in our experiments, with some additional 
modifications to construct both long-term and short-term context memories.
At any given time step we are given as input the history of the current conversation:
messages from the user $c^u_i$ at time step $i$ and the corresponding responses  $c^r_i$
from the model itself
 at the corresponding time steps, $i=1,\dots,t-1$. At the current time 
 $t$ we are only given
the input $c^u_t$ and the model has to respond.

\paragraph{Retrieving long-term memories} For each word in the last $N$ messages we perform a hash lookup to return all long-term memories (sentences) from a database that also contain that word.
Words above a certain frequency cutoff can be ignored to avoid sentences that only share
syntax or unimportant words. We employ the movie knowledge base  of
Sec. \ref{sec-qa} for our long-term memories, but potentially any text dataset could be used. See Figure \ref{memn2n-explain} for an example of this process.

\paragraph{Attention over memories} The sentences  $h_j$, ${\small j=1,\dots,H}$  returned from the hashing step plus the messages from the current conversation form the memory of the Memory Network\footnote{We also add time features to each memory to denote their position following \citep{sukhbaatar2015end}.}: 
\[
 x = (c^u_1,\dots,c^u_{t-1}, c^r_1, \dots, c^r_{t-1},h_1, \dots, h_H).
\]
The last user input $c^u_{t}$ is embedded using a matrix $A$ of size $d \times V$ where $d$ is the embedding
dimension and $V$ is the size of the vocabulary, giving $u=A c^u_{t}$.
 Each memory $x_i$ is embedded using the same matrix, giving $m_i = A x_i$.
The match between the input and the memories is then computed by taking the inner product followed by a softmax:
$p_i = \text{Softmax}(u^\top m_i)$ giving a probability vector over the memories.
The output memory representation is then constructed with $o = R \sum_i p_i m_i$ where $R$ is a $d \times d$
rotation matrix\footnote{Optionally, different dictionaries can be used for inputs, memories 
and outputs instead of being shared.}.
The memory output is then added to the original input $q=o + c^u_{t}$.
This procedure can then be stacked in what is called multiple ``hops'' of attention over the memory.

\paragraph{Generating the final prediction} The final prediction is then defined as:
$\hat{a} = \text{Softmax}(q^\top W y_1, \dots, q^\top W y_C)$ where there are $C$ candidate responses in $y$,
and $W$ is of dimension $V \times d$. For Tasks 1-3 the candidates are the set of words in the vocabulary, which 
are ranked for final evaluation, whereas for Task 4 the  candidates are target respones (sentences).

The whole model is trained using stochastic gradient descent  
by minimizing a standard cross-entropy loss between $\hat{a}$ and the true label $a$.

\begin{table}[h!]
\newcommand{\mc}[1]{\multicolumn{1}{l|}{#1}}
  \begin{center}
    \resizebox{0.95\linewidth}{!}{
        \begin{tabular}{l|l}
\hline
\textcolor{dgreen}{Long-Term} &\textcolor{blue}{\underline{Shaolin Soccer}} directed\_by \textcolor{blue}{\underline{Stephen Chow}}  \\
\textcolor{dgreen}{Memories~~~ $h_i$}   &\textcolor{blue}{\underline{Shaolin Soccer}} written\_by \textcolor{blue}{\underline{Stephen Chow}}   \\   
&\textcolor{blue}{\underline{Shaolin Soccer}} starred\_actors \textcolor{blue}{\underline{Stephen Chow}} \\ 
&\textcolor{blue}{\underline{Shaolin Soccer}} release\_year 2001        \\ 
&\textcolor{blue}{\underline{Shaolin Soccer}} has\_genre comedy         \\ 
&\textcolor{blue}{\underline{Shaolin Soccer}} has\_tags martial arts, kung fu soccer, \textcolor{blue}{\underline{stephen chow}} \\  
&\textcolor{blue}{\underline{Kung Fu Hustle}} directed\_by \textcolor{blue}{\underline{Stephen Chow}}  \\
&\textcolor{blue}{\underline{Kung Fu Hustle}} written\_by \textcolor{blue}{\underline{Stephen Chow}}    \\
&\textcolor{blue}{\underline{Kung Fu Hustle}} starred\_actors \textcolor{blue}{\underline{Stephen Chow}}  \\       
&\textcolor{blue}{\underline{Kung Fu Hustle}} has\_genre comedy action   \\
&\textcolor{blue}{\underline{Kung Fu Hustle}} has\_imdb\_votes famous     \\
&\textcolor{blue}{\underline{Kung Fu Hustle}} has\_tags comedy, action, martial arts, kung fu, china, soccer, hong kong, \textcolor{blue}{\underline{stephen chow}}\\ 
&The God of Cookery directed\_by \textcolor{blue}{\underline{Stephen Chow}} \\      
&The God of Cookery written\_by \textcolor{blue}{\underline{Stephen Chow}}    \\    
&The God of Cookery starred\_actors \textcolor{blue}{\underline{Stephen Chow}}  \\  
&The God of Cookery has\_tags hong kong \textcolor{blue}{\underline{Stephen Chow}}\\
&From Beijing with Love directed\_by \textcolor{blue}{\underline{Stephen Chow}}    \\
&From Beijing with Love written\_by \textcolor{blue}{\underline{Stephen Chow}}     \\
&From Beijing with Love starred\_actors \textcolor{blue}{\underline{Stephen Chow}}, Anita Yuen \\
 & ~~~~~~~~~~~~~~~~~ \textcolor{dgreen}{\dots $<$and more$>$ \dots} \\[0.5em]
\textcolor{dgreen}{Short-Term~~$c_1^u$}& 1) I'm looking a fun comedy to watch tonight, any ideas?\\
\textcolor{dgreen}{Memories~~~~$c_1^r$}&  2) Have you seen \textcolor{blue}{\underline{Shaolin Soccer}}? That was zany and great.. really funny but in a whacky way. \\[0.5em]
\textcolor{dgreen}{Input~~~~~~~~~~~~$c_2^u$}& 3) Yes! \textcolor{blue}{\underline{Shaolin Soccer}} and \textcolor{blue}{\underline{Kung Fu Hustle}} are so good I really need to find some more \textcolor{blue}{\underline{Stephen Chow}} \\
& films I feel like there is more awesomeness out there that I haven't discovered yet ...  \\[0.5em]
\textcolor{dgreen}{Output~~~~~~~~~~$y$}& \textcolor{dred}{4) God of Cookery is pretty great, one of his mid 90's hong kong martial art comedies.}   \\
          \hline
        \end{tabular}
      }
    \caption{\label{memn2n-explain} {\bf Memory Network long-term and short-term memories.} Blue underlined text indicates those words that hashed into the knowledge base to recall sentences from the long-term memory. Those, along with the recent short-term context (lines labeled 1 and 2) are used as input memories to the Memory Network along with the input (labeled 3). The desired goal is to output dialog line 4.}
  \end{center}
  \vspace*{-4ex}
\end{table}

\subsection{Supervised Embedding Models}\label{sem}

While one of the major uses of word embedding models is to learn
unsupervised embeddings over large unlabeled datasets such as in {\em Word2Vec} \citep{mikolov2013efficient}
there are also very effective word embedding models for training supervised models
when labeled data is available. The simplest approach which works suprisingly well is to sum the word embeddings
 of the input and the target independently and then compare them with a similarity metric such as inner product
or cosine similarity. A ranking loss is used to ensure the correct targets are
 ranked higher than any other targets.  Several variants of this approach exist.
For matching two documents supervised semantic indexing (SSI) was shown to be superior to unsupervised
latent semantic indexing (LSI) \citep{bai2009supervised}.
Similar methods were shown to outperform SVD for recommendation \citep{weston2013learning}.
However, we do not expect this method to work as well on question answering tasks,
as all the memorization must occur in the individual word embeddings, which was shown to perform 
poorly in \citep{bordes2014question}.
For example, consider asking the question ``who was born in Paris?'' and requiring
the word embedding for Paris to effectively contain all the pertinent information. However, for 
rarer items requiring less storage, performance may not be as degraded.
In general we believe this is a surprisingly strong baseline that is often neglected in evaluations.
Our implementation corresponds to a Memory Network with no attention over memory.

\subsection{Recurrent Language Models}

Recurrent Neural Networks (RNNs)
have proven successful at several tasks involving natural language, 
 language modeling \citep{mikolov2011extensions}, 
and have been applied recently to dialog \citep{sordoni2015neural,vinyals2015neural,shang2015neural}.
LSTMs are not known however for tasks such as QA  or item recommendation,
and so we expect them to find our datasets challenging.

There are a large number of variants of RNNs,  including
 Long-Short Term Memory activation units (LSTMs) \citep{hochreiter1997long},
 bidirectional LSTMs \citep{graves2012supervised}, {\em seq2seq} models \citep{sutskever2014sequence}, 
RNNs that take into account the document context \citep{mikolov2012context}
and RNNs that perform attention over their input in various different ways
 \citep{bahdanau2014neural,nips15_hermann,rush2015neural}.
Evaluating all these variants is beyond the scope of this work and we
instead use standard LSTMs as our baseline method\footnote{We used the code available at: \url{https://github.com/facebook/SCRNNs}}. 
However, we note that LSTMs with attention have
many properties in common with Memory Networks if the attention is applied over the same memory setup.

\subsection{Question Answering Systems}

For the particular case of Task 1 we can apply existing question answering systems.
There has been a recent surge in interest in such 
 systems that try to answer a question posed in
natural language by converting it into a database search over a knowledge base
\citep{berant2014semantic,kwiatkowski-EtAl:2013:EMNLP,fader2014open}, which is a setup natural for our QA task also.
However, such systems cannot easily solve any of our other tasks, for example our recommendation Task 2
does not involve looking up a factoid answer in a database.
Nevertheless, this allows us to compare the performance of end-to-end systems performant on all our tasks
to a standard QA benchmark.
We chose the method of \cite{bordes2014question}\footnote{We used the
  `{\em Path Representation}' for the knowledge base, as described in Sec.
  3.1 of \citet{bordes2014question}.} as our baseline. This system learns embeddings
that match questions to database entries, and then ranks the set
of entries, and has been shown to achieve
good performance on the {\sc WebQuestions} benchmark \citep{berant2013semantic}.

\subsection{Singular Value Decomposition}

Singular Value Decomposition (SVD) is a standard benchmark for recommendation,
being at the core of the best ensemble results in the Netflix challenge, see \cite{koren2011advances} for a review.
However, it has been shown to be outperformed by other flavors of matrix factorization,
in particular by using a ranking loss rather than squared loss \citep{weston2013learning} which we will
compare to (cf. sec \ref{sem}), as well as improvements like SVD++ \citep{koren2008factorization}.
Collaborative filtering methods are applicable to Task 2, but cannot easily be used for any of the other tasks.
Even for Task 2, while our dialog models use textual input, as shown in Table \ref{taskset2}, SVD requires 
a user $\times$ item matrix, so for this baseline we preprocessed the text to assign each entity an ID, and
throw away all other text. In contrast, 
the end-to-end dialog models have to learn to process the text as part of the task.  

\subsection{Information Retrieval Models}

To select candidate responses a standard baseline is nearest neighbour
 information retrieval (IR) \citep{isbell2000cobot,jafarpour2010filter,ritter2011data,sordoni2015neural}.
Two simple variants are often tried: given an input message, either
(i) find the most similar message in the (training) dataset and output the response from that exchange;
or (ii) find the most similar response to the input directly.
In both cases the standard measure of similarity is tf-idf weighted cosine similarity between the bags of words.
Note that that the Supervised Embedding Models of  Sec. \ref{sem} effectively implement the same kind of model 
(ii) but with a {\em learnt} similarity measure.
It has been shown previously that method (ii) performs better \citep{ritter2011data},
 and our initial IR experiments showed the same result.
%
Note that while (non-learning) IR systems can also be applied to other
tasks such as QA 
\citep{kolomiyets2011survey}
they require significant tuning to do so. Here we stick to a vanilla vector space 
model and hence only apply an IR baseline to Task 4.

\section{Results}

Our main results across all the models and tasks are given in Table 4.
Supervised Embeddings and Memory Networks are tested in two 
settings: trained and tested on all tasks separately,
or jointly on the combined Task 5.
Other methods are only evaluated on independent tasks.
In all cases, parameter search was performed on the development sets; parameter choices
are provided in the appendix.

\begin{table}[t]
\newcommand{\mc}[1]{\multicolumn{1}{l|}{#1}}
  \begin{center}
    \resizebox{1\linewidth}{!}{
      {\sc
        \begin{tabular}{l|cccc}
                     & QA Task & Recs Task  & QA+Recs Task  & Reddit Task \\
         \mc{Methods} & (hits@1) & (hits@100) & (hits@10) &  (hits@10)
          \\
          \hline
  {\small QA System  \citep{bordes2014question} }
                                    &  90.7   &    {\sc\small n/a}     &      {\sc\small n/a}     &   {\sc\small n/a}    \\

          \mc{SVD}                    &  {\sc\small n/a}    &    19.2    &      {\sc\small n/a}     &   {\sc\small n/a}  \\
          \mc{IR}                     &  {\sc\small n/a}    &   {\sc\small n/a}     &      {\sc\small n/a}     &   23.7   \\
          \hline
          LSTM                        &  ~6.5   &    27.1    &      19.9   &    11.8     \\
          \mc{Supervised Embeddings}  &  50.9   &    29.2    &      65.9    &   27.6   \\
          MemN2N                      &  79.3   &    28.6    &      81.7    &   29.2   \\
          \hline
          \hline
         \mc{Joint Supervised Embeddings }  &  43.6   &    28.1    &     58.9    &   14.5  \\
         \mc{Joint MemN2N                }  &  83.5   &    26.5    &     78.9    &   26.6  \\
        \end{tabular}
      }
   }
    \caption{\label{tab:full_res} {\bf Test results across all tasks.} Of those methods tested, supervised embeddings, LSTMs and MemN2N are easily applicable to all tasks. The other methods are standard benchmarks for individual tasks.
The final two rows are models trained on the Combined Task, of all tasks at once.
Evaluation uses the hits@k metric 
 (in percent)  with the value of $k$ given in the second row. 
}
  \end{center}
  \vspace*{-4ex}
\end{table}

\paragraph{Answering Factual Questions} 
Memory Networks and the baseline QA system are the
 two methods that have an explicit long-term memory via access to the knowledge base (KB).
On the task of answering factual questions where the answers are contained in the KB,
 they outperform the other methods convincingly, with LSTMS being particularly poor.
The latter is not unexpected as that method is good at language modeling, not question answering,
see e.g. \citet{weston2015towards}.
The baseline QA system, which is designed for this task, is superior to Memory Networks, indicating
there is still room for improvement in that model.
On the other hand, the latter's much more general
design allows it to perform well on our other dialog tasks, whereas the former
 is task specific.

\paragraph{Making Recommendations} In this task a long-term memory does not bring any improvement, with
LSTMs, Supervised Embeddings and Memory Networks all performing similarly, and all outperforming
the SVD baseline. Here, we conjecture LSTMs can perform well because it looks much more like a language modeling task, i.e. the input is a sequence of similar recommendations. 

\paragraph{Using Dialog History} In both  QA+Recommendations (Task 3) and Reddit (Task 4) 
Memory Networks outperform Supervised Embeddings due to their better use of context.
This can be seen  by breaking down the results by length of context: in
the first response they perform similarly, but Memory Networks show a relative improvement
on the second and third responses, see Tables \ref{tab:task3_res} and
\ref{tab:task4_res} in the appendix.
Note that these improvements come from the short term memory (dialog history), not from
the use of the KB, as we show Memory Networks results {\em without}
access to the KB and they perform similarly. We believe the QA performance in these cases
is not hindered by the lack of a KB because we ask questions based on fewer relations than in
Task 1 and it is easier to store the knowledge directly in the word embeddings.
The baseline IR model in Task 4 benefits from context too, it is compared with and without in
Table \ref{tab:task4_res}. 
LSTMs perform poorly: the posts in Reddit are quite long and the memory of the LSTM
is relatively short, as pointed out by  \citet{sordoni2015neural}. In that work they employed a linear
reranker that used LSTM prediction as features to better effect.
Testing more powerful recurrent networks such as 
 LSTMs with attention 
 on these  benchmarks remains as future work 
(although the latter is related to Memory Networks, which we do report).


\paragraph{Joint Learning} 
A truly end-to-end dialog system has to be good at all the skills in Tasks 1-4
(and more besides, i.e. this is necessary, but not sufficient).
We thus report results on our Combined Task for 
Supervised Embeddings and Memory Networks.
Supervised Embeddings still have the same failings as before on Tasks 1 and 3, but now
seem to perform even more poorly due to the difficulty of encoding all the necessary skills
in the word embeddings, so e.g., they now do significantly worse on Task 4. This is despite
us trying word embeddings of up to 2000 dimensions. Memory Networks fare
better, having only a slight loss in performance on Tasks 2-4 and a slight gain in Task 1.
In their case, the modeling power is not only in the word embeddings, but also
in the attention over the long-term and short-term memory,
 so it does not need as much capacity in the word embeddings.
However, the best achievable models would presumably have some {\em improvement} from training
across all the tasks, not a loss, and would perform at least as well as all the individual task
baselines (i.e. in this case, perform better at Task 1).

\section{Ubuntu Dialogue Corpus Results}
\label{sec:ubuntu}

\begin{table}[t]
  \begin{center}
    {\small 
    {\sc
      \begin{tabular}{ll|cc}
        & & Validation & Test \\
         Methods & & (hits@1) & (hits@1) \\
          \hline
          IR$^\dagger$ & &  n/a & 48.81 \\
          RNN$^\dagger$ & & n/a & 37.91  \\
          LSTM$^\dagger$ & & n/a & 55.22 \\
          \hline
          MemN2N & 1-hop & 57.23 & 56.25 \\
          MemN2N & 2-hops & 64.28 & 63.51 \\
          MemN2N & 3-hops & 64.31 & 63.72 \\
          MemN2N & 4-hops & 64.01 & 62.82 \\
          \hline
        \end{tabular}
      }
   }
    \caption{\label{tab:ubuntu_res}{\bf Ubuntu Dialog Corpus results.}
      The evaluation is retrieval-based, similar to that of Reddit
      (Task 4). For each dialog, the correct answer is mixed among 10
      random candidates; Hits@1 (in \%) are reported. Methods with $^\dagger$ have been ran
      by \citet{lowe2015ubuntu}.}
  \end{center}
  \vspace*{-4ex}
\end{table}

As no other authors have yet published results on our new benchmark,
to validate the quality of our results we also apply our best
performing model in other conditions by comparing it on the Ubuntu Dialog Corpus
\citep{lowe2015ubuntu}. 
In particular, this also allows us to compare to more sophisticated 
LSTMs models that are trained discriminatively using metric
learning, as well as additional baseline methods all trained by the authors. 
%
The Ubuntu Dialog Corpus contains almost 1M dialogs of more than 7
turns on average (900k dialogs for training, 20k for validation and 20k for
testing), and 100M million words. The corpus was scraped from the
Ubuntu IRC channel logs where users ask questions about
issues they are having with Ubuntu and get answers by
other users. Most chats can involve more than two users but a series
of heuristics to disentangle them into dyadic dialogs was used.

The evaluation is similar to that of Reddit (Task 4): each correct answer
has to be retrieved among a set of 10, mixed with 9 randomly chosen candidate
utterances. We report the Hits@1 in Table~\ref{tab:ubuntu_res}.\footnote{Results for
  the baselines from \citep{lowe2015ubuntu} differ to that from the v3
  of the arxiv paper, because the corpus has been updated since
  then. All results in Table~\ref{tab:ubuntu_res} use the latest
  version of the corpus.}
We used the same MemN2N architecture as before.
all models were selected using validation accuracy.
On this dataset, which has longer dialogs than those from the Movie
Dialog Corpus, we can see that running more hops on the memory with
the MemN2N improves performance: the 1-hop model performs similarly to
the LSTM but with 2-hops and more we can gain more than a +8\%
increase over the previous best reported model. Using even more hops
still improves over 1-hop but not much over 2-hops.

\vspace*{-1ex}
\section{Conclusion}
\vspace*{-1ex}

We have presented a new set of benchmark tasks designed to evaluate end-to-end dialog systems.
The movie dialog dataset measures how well such models can perform
at both goal driven dialog, of both objective and subjective goals thanks to 
evaluation metrics on question answering and recommendation tasks,  and at
  less goal driven {chit-chat}.
A true end-to-end model should perform
well at all these tasks, being a necessary but not sufficient condition for
a fully functional dialog agent.

We showed that some end-to-end neural networks models can perform reasonably across all tasks
compared to standard per-task baselines.
 Specifically, Memory Networks that incorporate short and long term memory can utilize
 local context and knowledge bases of facts to boost performance.
%
%
We believe this is promising because we showed these same
architectures also perform well on a separate dialog task, the Ubuntu Dialog Corpus,
and have been shown previously to work well on the
synthetic but challenging
bAbI tasks of \citet{Weston15}, 
and have no special engineering for the tasks or domain.
However, some limitations remain, in particular they do not perform as well as stand-alone 
QA systems for QA, and performance is also degraded rather than improved 
when training on all four tasks at once. Future work should try to overcome these problems.

 While our dataset focused on movies, there is nothing specific to
the task design which could not be transferred immediately to other domains, for example
sports, music, restaurants, and so on. Future work should create new tasks
 in this and other domains to ensure that models are firstly not overtuned for these goals,
and secondly to test further skills -- 
and to motivate the development of algorithms to be skillful at them.




\bibliography{iclr2016_conference}

\begin{thebibliography}{46}
\providecommand{\natexlab}[1]{#1}
\providecommand{\url}[1]{\texttt{#1}}
\expandafter\ifx\csname urlstyle\endcsname\relax
  \providecommand{\doi}[1]{doi: #1}\else
  \providecommand{\doi}{doi: \begingroup \urlstyle{rm}\Url}\fi

\bibitem[Bahdanau et~al.(2015)Bahdanau, Cho, and Bengio]{bahdanau2014neural}
Bahdanau, Dzmitry, Cho, Kyunghyun, and Bengio, Yoshua.
\newblock Neural machine translation by jointly learning to align and
  translate.
\newblock \emph{ICLR 2015}, 2015.

\bibitem[Bai et~al.(2009)Bai, Weston, Grangier, Collobert, Sadamasa, Qi,
  Chapelle, and Weinberger]{bai2009supervised}
Bai, Bing, Weston, Jason, Grangier, David, Collobert, Ronan, Sadamasa,
  Kunihiko, Qi, Yanjun, Chapelle, Olivier, and Weinberger, Kilian.
\newblock Supervised semantic indexing.
\newblock In \emph{Proceedings of the 18th ACM conference on Information and
  knowledge management}, pp.\  187--196. ACM, 2009.

\bibitem[Berant \& Liang(2014)Berant and Liang]{berant2014semantic}
Berant, Jonathan and Liang, Percy.
\newblock Semantic parsing via paraphrasing.
\newblock In \emph{Proceedings of the 52nd Annual Meeting of the Association
  for Computational Linguistics (ACL'14)}, Baltimore, USA, 2014.

\bibitem[Berant et~al.(2013)Berant, Chou, Frostig, and
  Liang]{berant2013semantic}
Berant, Jonathan, Chou, Andrew, Frostig, Roy, and Liang, Percy.
\newblock Semantic parsing on freebase from question-answer pairs.
\newblock In \emph{EMNLP}, pp.\  1533--1544, 2013.

\bibitem[Bordes et~al.(2014)Bordes, Chopra, and Weston]{bordes2014question}
Bordes, Antoine, Chopra, Sumit, and Weston, Jason.
\newblock Question answering with subgraph embeddings.
\newblock In \emph{Proc. EMNLP}, 2014.

\bibitem[Bordes et~al.(2015)Bordes, Usunier, Chopra, and
  Weston]{bordes2015large}
Bordes, Antoine, Usunier, Nicolas, Chopra, Sumit, and Weston, Jason.
\newblock Large-scale simple question answering with memory networks.
\newblock \emph{arXiv preprint arXiv:1506.02075}, 2015.

\bibitem[Cremonesi et~al.(2010)Cremonesi, Koren, and
  Turrin]{cremonesi2010performance}
Cremonesi, Paolo, Koren, Yehuda, and Turrin, Roberto.
\newblock Performance of recommender algorithms on top-n recommendation tasks.
\newblock In \emph{Proceedings of the fourth ACM conference on Recommender
  systems}, pp.\  39--46. ACM, 2010.

\bibitem[Fader et~al.(2014)Fader, Zettlemoyer, and Etzioni]{fader2014open}
Fader, Anthony, Zettlemoyer, Luke, and Etzioni, Oren.
\newblock Open question answering over curated and extracted knowledge bases.
\newblock In \emph{Proceedings of 20th SIGKDD Conference on Knowledge Discovery
  and Data Mining (KDD'14)}, New York City, USA, 2014. ACM.

\bibitem[Graves et~al.(2012)]{graves2012supervised}
Graves, Alex et~al.
\newblock \emph{Supervised sequence labelling with recurrent neural networks},
  volume 385.
\newblock Springer, 2012.

\bibitem[Griol et~al.(2008)Griol, Hurtado, Segarra, and
  Sanchis]{griol2008statistical}
Griol, David, Hurtado, Llu{\'\i}s~F, Segarra, Encarna, and Sanchis, Emilio.
\newblock A statistical approach to spoken dialog systems design and
  evaluation.
\newblock \emph{Speech Communication}, 50\penalty0 (8):\penalty0 666--682,
  2008.

\bibitem[Henderson(2015)]{henderson2015machine}
Henderson, Matthew.
\newblock Machine learning for dialog state tracking: A review.
\newblock In \emph{Proceedings of The First International Workshop on Machine
  Learning in Spoken Language Processing}, 2015.

\bibitem[Henderson et~al.(2014)Henderson, Thomson, and
  Williams]{henderson2014second}
Henderson, Matthew, Thomson, Blaise, and Williams, Jason.
\newblock The second dialog state tracking challenge.
\newblock In \emph{15th Annual Meeting of the Special Interest Group on
  Discourse and Dialogue}, pp.\  263, 2014.

\bibitem[Hermann et~al.(2015)Hermann, Ko\v{c}isk\'y, Grefenstette, Espeholt,
  Kay, Suleyman, and Blunsom]{nips15_hermann}
Hermann, Karl~Moritz, Ko\v{c}isk\'y, Tom\'a\v{s}, Grefenstette, Edward,
  Espeholt, Lasse, Kay, Will, Suleyman, Mustafa, and Blunsom, Phil.
\newblock Teaching machines to read and comprehend.
\newblock In \emph{Advances in Neural Information Processing Systems (NIPS)},
  2015.
\newblock URL \url{http://arxiv.org/abs/1506.03340}.

\bibitem[Hill et~al.(2015)Hill, Bordes, Chopra, and Weston]{hill2015goldilocks}
Hill, Felix, Bordes, Antoine, Chopra, Sumit, and Weston, Jason.
\newblock The goldilocks principle: Reading children's books with explicit
  memory representations.
\newblock \emph{arXiv preprint arXiv:1511.02301}, 2015.

\bibitem[Hirschman et~al.(1990)Hirschman, Dahl, McKay, Norton, and
  Linebarger]{hirschman1990beyond}
Hirschman, Lynette, Dahl, Deborah~A, McKay, Donald~P, Norton, Lewis~M, and
  Linebarger, Marcia~C.
\newblock Beyond class a: A proposal for automatic evaluation of discourse.
\newblock Technical report, DTIC Document, 1990.

\bibitem[Hochreiter \& Schmidhuber(1997)Hochreiter and
  Schmidhuber]{hochreiter1997long}
Hochreiter, Sepp and Schmidhuber, J{\"u}rgen.
\newblock Long short-term memory.
\newblock \emph{Neural computation}, 9\penalty0 (8):\penalty0 1735--1780, 1997.

\bibitem[Isbell et~al.(2000)Isbell, Kearns, Kormann, Singh, and
  Stone]{isbell2000cobot}
Isbell, Charles~Lee, Kearns, Michael, Kormann, Dave, Singh, Satinder, and
  Stone, Peter.
\newblock Cobot in lambdamoo: A social statistics agent.
\newblock In \emph{AAAI/IAAI}, pp.\  36--41, 2000.

\bibitem[Jafarpour et~al.(2010)Jafarpour, Burges, and
  Ritter]{jafarpour2010filter}
Jafarpour, Sina, Burges, Christopher~JC, and Ritter, Alan.
\newblock Filter, rank, and transfer the knowledge: Learning to chat.
\newblock \emph{Advances in Ranking}, 10, 2010.

\bibitem[Joulin \& Mikolov(2015)Joulin and Mikolov]{Joulin15}
Joulin, Armand and Mikolov, Tomas.
\newblock Inferring algorithmic patterns with stack-augmented recurrent nets.
\newblock \emph{arXiv preprint: 1503.01007}, 2015.

\bibitem[Kolomiyets \& Moens(2011)Kolomiyets and Moens]{kolomiyets2011survey}
Kolomiyets, Oleksandr and Moens, Marie-Francine.
\newblock A survey on question answering technology from an information
  retrieval perspective.
\newblock \emph{Information Sciences}, 181\penalty0 (24):\penalty0 5412--5434,
  2011.

\bibitem[Koren(2008)]{koren2008factorization}
Koren, Yehuda.
\newblock Factorization meets the neighborhood: a multifaceted collaborative
  filtering model.
\newblock In \emph{Proceedings of the 14th ACM SIGKDD international conference
  on Knowledge discovery and data mining}, pp.\  426--434. ACM, 2008.

\bibitem[Koren \& Bell(2011)Koren and Bell]{koren2011advances}
Koren, Yehuda and Bell, Robert.
\newblock Advances in collaborative filtering.
\newblock In \emph{Recommender systems handbook}, pp.\  145--186. Springer,
  2011.

\bibitem[Kwiatkowski et~al.(2013)Kwiatkowski, Choi, Artzi, and
  Zettlemoyer]{kwiatkowski-EtAl:2013:EMNLP}
Kwiatkowski, Tom, Choi, Eunsol, Artzi, Yoav, and Zettlemoyer, Luke.
\newblock Scaling semantic parsers with on-the-fly ontology matching.
\newblock In \emph{Proceedings of the 2013 Conference on Empirical Methods in
  Natural Language Processing (EMNLP'13)}, Seattle, USA, October 2013.

\bibitem[Lowe et~al.(2015)Lowe, Pow, Serban, and Pineau]{lowe2015ubuntu}
Lowe, Ryan, Pow, Nissan, Serban, Iulian, and Pineau, Joelle.
\newblock The ubuntu dialogue corpus: A large dataset for research in
  unstructured multi-turn dialogue systems.
\newblock \emph{arXiv preprint arXiv:1506.08909}, 2015.

\bibitem[Mikolov \& Zweig(2012)Mikolov and Zweig]{mikolov2012context}
Mikolov, Tomas and Zweig, Geoffrey.
\newblock Context dependent recurrent neural network language model.
\newblock In \emph{SLT}, pp.\  234--239, 2012.

\bibitem[Mikolov et~al.(2011)Mikolov, Kombrink, Burget, {\v{C}}ernock{\`y}, and
  Khudanpur]{mikolov2011extensions}
Mikolov, Tom{\'a}{\v{s}}, Kombrink, Stefan, Burget, Luk{\'a}{\v{s}},
  {\v{C}}ernock{\`y}, Jan~Honza, and Khudanpur, Sanjeev.
\newblock Extensions of recurrent neural network language model.
\newblock In \emph{Acoustics, Speech and Signal Processing (ICASSP), 2011 IEEE
  International Conference on}, pp.\  5528--5531. IEEE, 2011.

\bibitem[Mikolov et~al.(2013)Mikolov, Chen, Corrado, and
  Dean]{mikolov2013efficient}
Mikolov, Tomas, Chen, Kai, Corrado, Greg, and Dean, Jeffrey.
\newblock Efficient estimation of word representations in vector space.
\newblock \emph{arXiv:1301.3781}, 2013.

\bibitem[Narasimhan et~al.(2015)Narasimhan, Kulkarni, and
  Barzilay]{narasimhan2015language}
Narasimhan, Karthik, Kulkarni, Tejas, and Barzilay, Regina.
\newblock Language understanding for text-based games using deep reinforcement
  learning.
\newblock \emph{arXiv preprint arXiv:1506.08941}, 2015.

\bibitem[Paek(2001)]{paek2001empirical}
Paek, Tim.
\newblock Empirical methods for evaluating dialog systems.
\newblock In \emph{Proceedings of the workshop on Evaluation for Language and
  Dialogue Systems-Volume 9}, pp.\ ~2. Association for Computational
  Linguistics, 2001.

\bibitem[Ramachandran et~al.(2014)Ramachandran, Yeh, Jarrold, Douglas,
  Ratnaparkhi, Provine, Mendel, and Emfield]{ramachandran2014end}
Ramachandran, Deepak, Yeh, Peter~Z, Jarrold, William, Douglas, Benjamin,
  Ratnaparkhi, Adwait, Provine, Ronald, Mendel, Jeremy, and Emfield, Adam.
\newblock An end-to-end dialog system for tv program discovery.
\newblock In \emph{Spoken Language Technology Workshop (SLT), 2014 IEEE}, pp.\
  602--607. IEEE, 2014.

\bibitem[Ritter et~al.(2011)Ritter, Cherry, and Dolan]{ritter2011data}
Ritter, Alan, Cherry, Colin, and Dolan, William~B.
\newblock Data-driven response generation in social media.
\newblock In \emph{Proceedings of the Conference on Empirical Methods in
  Natural Language Processing}, pp.\  583--593. Association for Computational
  Linguistics, 2011.

\bibitem[Rojas-Barahona et~al.(2012)Rojas-Barahona, Lorenzo, and
  Gardent]{rojas2012end}
Rojas-Barahona, Lina~M, Lorenzo, Alejandra, and Gardent, Claire.
\newblock An end-to-end evaluation of two situated dialog systems.
\newblock In \emph{Proceedings of the 13th Annual Meeting of the Special
  Interest Group on Discourse and Dialogue}, pp.\  10--19. Association for
  Computational Linguistics, 2012.

\bibitem[Rush et~al.(2015)Rush, Chopra, and Weston]{rush2015neural}
Rush, Alexander~M, Chopra, Sumit, and Weston, Jason.
\newblock A neural attention model for abstractive sentence summarization.
\newblock \emph{Proceedings of EMNLP}, 2015.

\bibitem[Shang et~al.(2015)Shang, Lu, and Li]{shang2015neural}
Shang, Lifeng, Lu, Zhengdong, and Li, Hang.
\newblock Neural responding machine for short-text conversation.
\newblock \emph{arXiv preprint arXiv:1503.02364}, 2015.

\bibitem[Sordoni et~al.(2015)Sordoni, Galley, Auli, Brockett, Ji, Mitchell,
  Nie, Gao, and Dolan]{sordoni2015neural}
Sordoni, Alessandro, Galley, Michel, Auli, Michael, Brockett, Chris, Ji,
  Yangfeng, Mitchell, Margaret, Nie, Jian-Yun, Gao, Jianfeng, and Dolan, Bill.
\newblock A neural network approach to context-sensitive generation of
  conversational responses.
\newblock \emph{Proceedings of NAACL}, 2015.

\bibitem[Sukhbaatar et~al.(2015)Sukhbaatar, Szlam, Weston, and
  Fergus]{sukhbaatar2015end}
Sukhbaatar, Sainbayar, Szlam, Arthur, Weston, Jason, and Fergus, Rob.
\newblock End-to-end memory networks.
\newblock \emph{Proceedings of NIPS}, 2015.

\bibitem[Sutskever et~al.(2014)Sutskever, Vinyals, and
  Le]{sutskever2014sequence}
Sutskever, Ilya, Vinyals, Oriol, and Le, Quoc~VV.
\newblock Sequence to sequence learning with neural networks.
\newblock In \emph{Advances in neural information processing systems}, pp.\
  3104--3112, 2014.

\bibitem[Vinyals \& Le(2015)Vinyals and Le]{vinyals2015neural}
Vinyals, Oriol and Le, Quoc.
\newblock A neural conversational model.
\newblock \emph{arXiv preprint arXiv:1506.05869}, 2015.

\bibitem[Walker et~al.(1997)Walker, Litman, Kamm, and
  Abella]{walker1997paradise}
Walker, Marilyn~A, Litman, Diane~J, Kamm, Candace~A, and Abella, Alicia.
\newblock Paradise: A framework for evaluating spoken dialogue agents.
\newblock In \emph{Proceedings of the eighth conference on European chapter of
  the Association for Computational Linguistics}, pp.\  271--280. Association
  for Computational Linguistics, 1997.

\bibitem[Walker et~al.(2003)Walker, Prasad, and Stent]{walker2003trainable}
Walker, Marilyn~A, Prasad, Rashmi, and Stent, Amanda.
\newblock A trainable generator for recommendations in multimodal dialog.
\newblock In \emph{INTERSPEECH}, 2003.

\bibitem[Weston et~al.(2015{\natexlab{a}})Weston, Bordes, Chopra, and
  Mikolov]{Weston15}
Weston, J., Bordes, A., Chopra, S., and Mikolov, T.
\newblock Towards {AI}-complete question answering: A set of prerequisite toy
  tasks.
\newblock \emph{arXiv preprint: 1502.05698}, 2015{\natexlab{a}}.

\bibitem[Weston et~al.(2013)Weston, Yee, and Weiss]{weston2013learning}
Weston, Jason, Yee, Hector, and Weiss, Ron~J.
\newblock Learning to rank recommendations with the k-order statistic loss.
\newblock In \emph{Proceedings of the 7th ACM conference on Recommender
  systems}, pp.\  245--248. ACM, 2013.

\bibitem[Weston et~al.(2015{\natexlab{b}})Weston, Bordes, Chopra, and
  Mikolov]{weston2015towards}
Weston, Jason, Bordes, Antoine, Chopra, Sumit, and Mikolov, Tomas.
\newblock Towards ai-complete question answering: a set of prerequisite toy
  tasks.
\newblock \emph{arXiv preprint arXiv:1502.05698}, 2015{\natexlab{b}}.

\bibitem[Weston et~al.(2015{\natexlab{c}})Weston, Chopra, and
  Bordes]{weston2014memory}
Weston, Jason, Chopra, Sumit, and Bordes, Antoine.
\newblock Memory networks.
\newblock \emph{Proceedings of ICLR}, 2015{\natexlab{c}}.

\bibitem[Whittaker et~al.(2002)Whittaker, Walker, and Moore]{whittaker2002fish}
Whittaker, Steve, Walker, Marilyn~A, and Moore, Johanna~D.
\newblock Fish or fowl: A wizard of oz evaluation of dialogue strategies in the
  restaurant domain.
\newblock In \emph{LREC}, 2002.

\bibitem[Williams et~al.(2013)Williams, Raux, Ramachandran, and
  Black]{williams2013dialog}
Williams, Jason, Raux, Antoine, Ramachandran, Deepak, and Black, Alan.
\newblock The dialog state tracking challenge.
\newblock In \emph{Proceedings of the SIGDIAL 2013 Conference}, pp.\  404--413,
  2013.

\end{thebibliography}
\bibliographystyle{iclr2016_conference}

\appendix
\newpage
\section{Further Experimental Details}

\paragraph{Dictionary}
For all models we built a dictionary using all the known entities in the KB
(e.g. ``Bruce Willis'' and ``Die Hard'' are single dictionary elements).
This allows us to output a single symbol for QA and Recommendation in order
to predict an entity, rather than having to construct the answer out of words,
making training and evaluation of the task simpler.
The rest of the dictionary is built of unigrams that are not covered by our entity
dictionary, where we removed other words (but not entities) 
with frequency less than 5. Overall this gives  a dictionary of size 189472,
which includes 75542 entities. All entries and texts were lower-cased.
Our text parser to convert to the dictionary representation 
is then very simple: it goes left to right,
consuming the largest $n$-gram at each step.

\vspace{-1mm}
\paragraph{Memory Networks}
For most of the tasks the optimal number of hops was 1, except for Task 3
where 2 or 3 hops outperform 1. See Table \ref{tab:task3_res} and the parameter
choices in Sec.  \ref{hyperparams}.
For the joint task (Task 5),  to achieve best performance 
we increased the capacity compared to the individual task models 
by using different dictionaries for the input,
memory and output layers, see Sec.  \ref{hyperparams}.
Additionally, we pre-trained the weights by training without
the long-term memory for speed.

\vspace{-1mm}
\paragraph{Supervised Embedding Models}
We tried two flavors of supervised embedding model:
(i) a model $f(x, y) = x^\top U^\top U y$  (``single dictionary model'');
and (ii) a model $f(x, y) = x^\top U^\top V y$  (``two dictionary model''). 
That is, the latter has two sets of word embeddings depending on whether the word
is in the input+context, or the label. The input and context are concatenated together
to form a bag of words in either case. It turns out method (i) works better on Tasks 1 and 4,
and method (ii) works better on Tasks 2 \& 3. Some of the reasons why that is so are easy to
understand: on Tasks 2  and 3 (recommendations)  a single dictionary model favors predicting 
the same movies that are already in the input context, which are never correct. However,
it appears that on Tasks 1 and 4 the two dictionary model appears to overfit to some degree.
This partially explains why the model
overall is worse on the joint dataset (Task 5). See Sec. \ref{hyperparams} for more details.

\vspace{-1mm}
\paragraph{LSTMs}
LSTMs performed poorly on Task 4 and we spent some time trying to improve these results.
Despite the perplexity looking reasonable ($\sim$96 on the training set, and $\sim$105 on the validation set) after training for $\sim$6 days, we still obtain poor results distinguishing between
candidates. 
We also tried {\em Seq2Seq} models (without attention or metric learning) 
and did not obtain improvements.
Part of the problem is that posts in Reddit vary from very short (a few words) to very long (several paragraphs) and one natural procedure to try -- 
computing the probability of those sequences seeded by the input --  gives very
unbalanced results, and tends to select the shorter ones, ending up with worse than random performance.
Further, computationally the whole procedure is then very slow compared to all other methods tested.
Memory Networks and supervised embeddings need
 to compute the inner product between embedded inputs and outputs, and hence the
 the candidates can be embedded once and cached  for the whole test set.
This trick is not applicable to the method described above rendering it much slower. 
To deal with the speed issue one can use our supervised embedding model as
a first step, and then only reranking the top 100 results with the LSTM to make it tractable, however
performance is still poor as mentioned. 
We obtained improved results by instead adopting the approach of \cite{narasimhan2015language}:
we take the representation for a dialog message as the average embedding over the hidden states
as the symbols are consumed (at each step of the recurrence).
We also note that \cite{lowe2015ubuntu} report
good results (on a different dataset, the Ubuntu Corpus) by training an additional metric learner
on top of an LSTM representation, which we have not tried.
However, we do compare that approach to Memory Networks on that corpus in Section \ref{sec:ubuntu}.

\vspace{-1mm}
\paragraph{Information Retrieval}
Aside from the models described in the main paper, we
 also experimented with a hybrid relevance feedback approach:
find the most similar message in the history, add the response to the query (with a certain weight) and
then score candidate responses with the combined input.
However, the relevance feedback model did not help: as we increase
the feedback parameter (how much to use the retrieved response) the model only degrades,
see Table \ref{tab:task4_res} for the performance adding with a weight of 0.5.

\section{Optimal hyper-parameter values} \label{hyperparams}

Hyperparameters of all learning models have been set using grid search on
the validation set. The main hyperparameters are embedding dimension $d$, learning rate $\lambda$,
number of dictionaries $w$, number of hops $K$ for MemNNs and unfolding depth  $\text{blen}$ for LSTMs.
All models are implemented in the Torch library (see \url{torch.ch}).

\paragraph{Task 1 (QA)}

\begin{itemize}
\item QA System of \cite{bordes2014question}: $\lambda=0.001$, $d=50$.

\item Supervised Embedding Model:  $\lambda=0.05$, $d=50$, $w=1$.

\item MemN2N:  $\lambda=0.005$, $d=50$, $w=1$, $K=1$.

\item LSTM:  $\lambda=0.001$, $d=100$, $\text{blen}=10$.

\end{itemize}

\paragraph{Task 2 (Recomendation)}

\begin{itemize}
\item SVD: $d=50$.

\item Supervised Embedding Model:  $\lambda=0.005$, $d=200$, $w=2$.

\item MemN2N:  $\lambda=0.01$, $d=1000$, $w=1$, $K=1$.

\item LSTM:  $\lambda=0.01$, $d=100$, $\text{blen}=10$.

\end{itemize}

\paragraph{Task 3 (QA+Recommendation)}

\begin{itemize}

\item Supervised Embedding Model:  $\lambda=0.005$, $d=1000$, $w=2$.

\item MemN2N:  $\lambda=0.001$, $d=50$, $w=1$, $K=3$.

\item LSTM:  $\lambda=0.001$, $d=100$, $\text{blen}=10$.

\end{itemize}

\paragraph{Task 4 (Reddit)}

\begin{itemize}
\item Supervised Embedding Model:  $\lambda=0.1$, $d=1000$, $w=1$.

\item MemN2N:  $\lambda=0.01$, $d=1000$, $w=1$, $K=1$.

\item LSTM:  $\lambda=0.01$, $d=512$, $\text{blen}=15$.

\end{itemize}

\paragraph{Joint Task}

We chose hyperparameters by taking the mean performance over the four tasks,
after scaling each task by the best performing model on that task on the development
set in order to normalize the metrics.

\begin{itemize}
\item Supervised Embedding Model:  $\lambda=0.01$, $d=1000$, $w=2$.

\item MemN2N:  $\lambda=0.005$, $d=1000$, $w=3$.

\end{itemize}

\paragraph{Ubuntu Dialog Corpus}

Hyperparameters of the MemN2N have been set using grid search on the
validation set. We report the best models with $K=1,\,2,\,3,\,4$ in
the paper; other hyperparameters were $\lambda=0.001$, $d=256$.

\newpage 

\section{Further Detailed Results}

\subsection{Breakdown of Task 1 (QA) results by question type}

\begin{table}[!htbp]
\newcommand{\mc}[1]{\multicolumn{1}{l|}{#1}}
  \begin{center}
    \resizebox{0.75\linewidth}{!}{
      {\sc
        \begin{tabular}{l|cccccc}
              & \multicolumn{2}{c}{\small QA System of} &  \multicolumn{2}{c}{Supervised}  &          \\
              &  \multicolumn{2}{c}{\small{\cite{bordes2014question}}} & \multicolumn{2}{c}{Embeddings} &     \multicolumn{2}{c}{MemN2N} \\
     Task     & h@1 & h@10 & h@1 & h@10 & h@1 & h@10 \\
\hline
writer to movie   & 98.7 & 98.7 & 77.3 & 90.8  &    77.6 &  95.5  \\
tag to movie      & 71.8 & 71.8 & 53.4 & 96.1 &     61.4 &  88.6  \\
movie to year     & 89.8 & 89.8 & ~~3.4 & 25.4  &   87.3 &  92.1  \\
movie to writer   & 88.8 & 89.5 & 61.7 & 93.6 &     73.5 &  84.1  \\
movie to tags     & 84.5 & 85.3 & 36.8 & 92.0  &    79.9 &  95.1  \\
movie to language & 94.6 & 94.8 & 45.2 & 84.7 &     90.1 &  97.6  \\
movie to genre    & 93.0 & 93.5 & 46.4 & 95.0 &     92.5 &  99.4  \\
movie to director & 88.2 & 88.2 & 52.3 & 90.1 &     78.3 &  87.1  \\
movie to actors   & 88.5 & 88.5 & 64.5 & 95.2 &     68.4 &  87.2  \\
director to movie & 98.3 & 98.3 & 61.4 & 93.8 &     71.5 &  91.0  \\
actor to  movie   & 98.9 & 98.9 & 79.0 & 89.4 &     76.7 &  96.7  \\
\hline
total             & 90.7 & 91.0  & 50.9 & 82.97 &     78.9 & 91.8 \\
       \end{tabular}
      }
   }
    \caption{\label{tab:task1_res} {\bf QA task test performance per question type} 
(h@1 / h@10 metrics).}
  \end{center}
\end{table}

\subsection{Breakdown of Task 3 (QA+Recommendation) results by response type}

\begin{table}[h!]
\newcommand{\mc}[1]{\multicolumn{1}{l|}{#1}}
  \begin{center}
    \resizebox{0.8\linewidth}{!}{
      {\sc
        \begin{tabular}{l|c|cccc}

        &            Whole  &  Response 1  & Response 2 & Response 3  \\
          \mc{Methods} &   Test Set &  (Recs) & (QA) &  (Similar) \\
         \hline
      \mc{Supervised Embeddings}  & 56.0 & 56.7 &  76.2 & 38.8 \\
      LSTM                        & 19.9 & 35.3 & 14.3 &  ~9.2 \\
      MemN2N (1 hop)              & 70.5 &  47.0 & 89.2 & 76.5 \\
      MemN2N (2 hops)             & 76.8 &  53.4 & 90.1 & 88.6 \\
      MemN2N (3 hops)             & 75.4 &  52.6 & 90.0 & 84.2 \\
      MemN2N (3 hops, -KB)        & 75.9 &  54.3 & 85.0 & 91.5 \\
         \hline
        \end{tabular}
      }
    }
    \caption{\label{tab:task3_res} {\bf QA+Recommendation task test results} (h@10 metric). The last row shows MemN2N without access to a long-term memory (KB).}
  \end{center}
\end{table}

\subsection{Breakdown of Task 4 (Reddit) results by response type}

\begin{table}[h!]
\newcommand{\mc}[1]{\multicolumn{1}{l|}{#1}}
  \begin{center}
    \resizebox{1\linewidth}{!}{
      {\sc
        \begin{tabular}{l|c|cccc}
 &         Whole & Entity & \\
          \mc{Methods} & 
         Test Set & Matched & Response 1 & Response 2 & Response 3+ \\
         \hline
         \mc{IR (query+context)}  &   23.7 & 49.0 & 21.1 & 26.4 & 30.0  \\
         \mc{IR (query)}          &   23.1 & 48.3 & 21.1 & 25.7 & 27.9  \\
         \mc{IR (query) RF=0.05}  &   19.2 & 40.8 & 18.3 & 21.2 & 21.4   \\
         \hline
      \mc{Supervised Embeddings}  &   27.6 & 54.1 & 24.8 & 30.4 & 33.1   \\ 
      MemN2N (-KB)                &  29.6   &  57.0  &  25.6  &  34.2 & 37.2  \\
      MemN2N                      &  29.2   &  56.4  &  25.4  &  32.9 & 37.0  \\
         \hline
        \end{tabular}
      }
    }
    \caption{\label{tab:task4_res} {\bf Reddit task test results} (h@10 metric).
 {\sc MemN2N (-KB)} is the Memory Network model without access to the knowledge base.
}
  \end{center}
\end{table}

\end{document}